\documentclass[10pt]{article}

\usepackage{graphicx}
\usepackage{amsmath}
\usepackage{amssymb}
\usepackage{hyperref}
\usepackage{geometry}
\usepackage[authoryear]{natbib}
\usepackage{algorithm}
\usepackage{algpseudocode}
\usepackage{subcaption}

\begin{document}

\title{Flash Window Attention: speedup the attention computation for Swin Transformer}
\author{
    Zhendong Zhang \\
    \texttt{zhd.zhang.ai@gmail.com}
    \vspace{0.0em}
}

\maketitle

\begin{abstract}
To address the high resolution of image pixels, the Swin Transformer introduces window attention. 
This mechanism divides an image into non-overlapping windows and restricts attention computation to within each window, 
significantly enhancing computational efficiency. To further optimize this process, 
one might consider replacing standard attention with flash attention, which has proven to be more efficient in language models. 
However, a direct substitution is ineffective. Flash attention is designed for long sequences, 
whereas window attention deals with shorter sequences but must handle numerous of them in parallel.
In this report, we present an optimized solution called Flash Window Attention, tailored specifically for window attention. 
Flash Window Attention improves attention computation efficiency by up to 300\% and 
enhances end-to-end runtime efficiency by up to 30\%.
Our code is available at \url{github.com/zzd1992/FlashWindowAttention}.
\end{abstract}

\section{Introduction}
The Transformer architecture \cite{Vaswani2017AttentionIA} has emerged as the dominant neural network model for sequence modeling. 
Its remarkable success in natural language processing has inspired researchers to adapt it for computer vision tasks. 
However, a key challenge in this adaptation lies in the high resolution of image pixels, 
as the computational complexity of attention mechanisms scales quadratically with the number of pixels.
To overcome this limitation, the Swin Transformer \cite{Liu2021SwinTH} introduces window attention. 
This approach computes attention locally within fixed-size, non-overlapping windows that partition the image. 
By limiting computations to these windows, the complexity of the attention mechanism becomes linear with respect to the number of pixels.
To further facilitate information exchange across windows, Swin Transformer employs a shifted window mechanism, 
where the image is shifted before applying the window partitioning. This is where the name Swin (Shifted window) comes from.

Flash attention \cite{Dao2022FlashAttentionFA, Dao2023FlashAttention2FA} is a widely adopted technique for enhancing the efficiency of attention computation, 
particularly in large language models (LLMs). Its core innovation lies in avoiding the storage of the attention matrix in GPU global memory, 
which can be a significant bottleneck.
To achieve this, flash attention processes the query, key, and value matrices in chunks along the sequence dimension. 
For each query chunk, the algorithm computes the attention matrix and a temporary output entirely within on-chip SRAM. 
It then iterates through the key and value chunks, updating the output for the same query chunk at each step. 
This process continues until all key and value chunks have been processed.
By leveraging the high-speed access of on-chip SRAM compared to global memory, 
flash attention dramatically improves the efficiency of attention computation for long sequences. 
For further details on the algorithm and implementation, refer to \cite{Dao2022FlashAttentionFA, Dao2023FlashAttention2FA}.

To further enhance the efficiency of window attention, one potential approach is to replace standard attention with flash attention. 
However, a direct replacement is ineffective. Flash attention is specifically optimized for long sequences by tiling along the sequence dimension, 
but window attention involves short sequences with numerous instances processed in parallel. 
For example, in the Swin Transformer, the sequence length is only 49, making sequence tiling ineffective.
In this report, we propose an optimized flash scheme tailored for short sequences, building on the following two key observations:
\begin{itemize}
    \item For short sequences, the entire attention matrix can be stored in on-chip SRAM, eliminating the need for slower global memory access.
    \item Attention computation can be decomposed along feature dimension.
\end{itemize}

Given a query/key/value pair, we split them into chunks along the feature dimension.
We first compute and accumulate the attention matrix on chip SRAM until all query/key chunks are visited. 
Then we compute the attention output using value chunk by chunk. This approach eliminates the need to store the attention matrix in global memory.
The query/key/value are divided into chunks to reduce the on chip SRAM usage. 
We call the proposed method Flash Window Attention. For forward pass, the global memories are accessed only once.

\section{Methodology}
\subsection{Problem Formulation}
For Swin Transformer, $\mathbf{Q/K/V}$ are represented as $H \times W \times C$ tensors, where $H$ is the height, $W$ is the width, 
and $C$ is the number of channels. Then they are rearanged by window partition as follows:
\begin{equation}
    H \times W \times C \rightarrow (\frac{H \times W}{k^2}) \times k^2 \times C = N \times L \times C
\end{equation}  
where $k$ the window size. Since the computation can be parallelized along the first dimension, we focus on 
$\mathbf{Q/K/V}$ matrices with shape $L \times C$. The attention output is computed as following three steps:
\begin{align}
    \mathbf{S} &= \mathbf{Q}\mathbf{K}^T \\
    \mathbf{P} &= \text{softmax}(\mathbf{S}) \\
    \mathbf{O} &= \mathbf{PV}
\end{align}

Following the spirit of flash attention, we want to avoid storing the matrix $\mathbf{S, P}$ in global memory. 
Since tiling along sequence dimension brings no benefit, we will use a different scheme.

\subsection{Tiling along Feature Dimension}
We split $\mathbf{Q/K/V}$ matrices into chunks along the feature dimension, i.e. $\mathbf{Q} = [\mathbf{Q}_1, \dots, \mathbf{Q}_r]$ 
where $\mathbf{Q}_i$ has shape $L \times C/r$. Then we accumulate the attention matrix $\mathbf{S}$ chunk by chunk:
\begin{equation}
    \mathbf{S} = \sum_{i=1}^r \mathbf{Q}_i \mathbf{K}_i^T
\end{equation}
We also compute the attention output $\mathbf{O}$ chunk by chunk:
\begin{equation}
    \mathbf{O}_i = \mathbf{P}\mathbf{V}_i
\end{equation}

\subsection{Implementation and Analysis}

\begin{algorithm}[t]
    \caption{Flash Window Attention forward}
    \label{alg:fwd}
    \begin{algorithmic}
    
    \Require $\mathbf{Q}, \mathbf{K}, \mathbf{V} \in \mathbb{R}^{L \times C}$ in global memory, number of chunks $r$
    \Ensure Attention output $\mathbf{O} \in \mathbb{R}^{L \times C}$ 
    
    \State Divide $\mathbf{Q}$ into $r$ chunks: $\mathbf{Q}_1, \dots, \mathbf{Q}_r$ of size $L \times C/r$ each
    \State Divide $\mathbf{K}$ into $r$ chunks: $\mathbf{K}_1, \dots, \mathbf{K}_r$ of size $L \times C/r$ each
    \State Divide $\mathbf{V}$ into $r$ chunks: $\mathbf{V}_1, \dots, \mathbf{V}_r$ of size $L \times C/r$ each
    
    \State On chip, initialize $\mathbf{S} \in \mathbb{R}^{L \times L}$ to zero
    \For{$i = 1$ to $r$}
        \State Load $\mathbf{Q}_i, \mathbf{K}_i$ from global memory to on-chip SRAM
        \State On chip, compute $\mathbf{S} = \mathbf{S} + \mathbf{Q}_i \mathbf{K}_i^T$
    \EndFor
    \State On chip, compute $\mathbf{P} = \text{softmax}(\mathbf{S})$
    \For{$i = 1$ to $r$}
        \State Load $\mathbf{V}_i$ from global memory to on-chip SRAM
        \State On chip, compute $\mathbf{O}_i = \mathbf{P} \mathbf{V}_i$
        \State Write $\mathbf{O}_i$ to global memory of $\mathbf{O}$
    \EndFor
    
    \State Return $\mathbf{O}$
    
\end{algorithmic}
\end{algorithm}

When $L$ is small, the entire attention matrix $\mathbf{S,P}$ can be stored on chip SRAM. 
This leads to the forward algorithm \ref{alg:fwd}. As we can see, the global memories of $\mathbf{Q,K,V,O}$ are accessed
only once. Thus, algorithm \ref{alg:fwd} minimizes the global memory access for forward pass. 
For on chip SRAM, the space complexity for storing the attention matrix is $\Theta(L^2)$, and the space complexity for storing 
query/key/value chunks is $\Theta(LC/r)$. Therefore, the space complexity is $\Theta(L^2+LC/r)$. More specifically,
the peak on chip memory usage is $L^2 + 2LC/r$. Because we need to store $\mathbf{S, Q_i, K_i}$ at the same time. 
Under typical settings such as $L=64$ and $C/r=16$, the peak on chip memory usage is 24kb for fp32 format.
This is well within the capacity of the L1 cache on modern GPUs. 
For instance, the NVIDIA GeForce RTX 4090 features an L1 cache of 128 KB per SM, making it sufficient for usage.

\begin{algorithm}[t]
    \caption{Flash Window Attention backward}
    \label{alg:bwd}
    \begin{algorithmic}
    
    \Require $\mathbf{Q}, \mathbf{K}, \mathbf{V}, \mathbf{dO} \in \mathbb{R}^{L \times C}$ in global memory, number of chunks $r$
    \Ensure $\mathbf{dQ}, \mathbf{dK}, \mathbf{dV} \in \mathbb{R}^{L \times C}$ 
    
    \State Divide $\mathbf{Q, K, V, dO}$ into $r$ chunks: each of size $L \times C/r$
    \State On chip, initialize $\mathbf{P, dP} \in \mathbb{R}^{L \times L}$ to zero
    \For{$i = 1$ to $r$}
        \State Load $\mathbf{Q}_i, \mathbf{K}_i$ from global memory to on-chip SRAM
        \State On chip, compute $\mathbf{P} = \mathbf{P} + \mathbf{Q}_i \mathbf{K}_i^T$
    \EndFor
    \State On chip, compute $\mathbf{P} = \text{softmax}(\mathbf{P})$
    \For{$i = 1$ to $r$}
        \State Load $\mathbf{dO}_i, \mathbf{V}_i$ from global memory to on-chip SRAM
        \State On chip, compute $\mathbf{dV}_i$ = $\mathbf{P}^T \mathbf{dO}_i$
        \State Write $\mathbf{dV}_i$ to global memory of $\mathbf{dV}$
        \State On chip, compute $\mathbf{dP} = \mathbf{dP} + \mathbf{dO}_i \mathbf{V}_i^T$
    \EndFor
    
    \State On chip, compute $\mathbf{dS} \in \mathbb{R}^{L \times L}$, where $dS_{ij} = P_{ij}(dP_{ij} - \sum_{l}P_{il}dP_{il})$
    
    \For{$i = 1$ to $r$}
        \State Load $\mathbf{Q}_i, \mathbf{K}_i$ from global memory to on-chip SRAM
        \State On chip, compute $\mathbf{dQ}_i$ = $\mathbf{dS} \mathbf{K}_i$
        \State Write $\mathbf{dQ}_i$ to global memory of $\mathbf{dQ}$
        \State On chip, compute $\mathbf{dK}_i$ = $\mathbf{dS}^T \mathbf{Q}_i$
        \State Write $\mathbf{dK}_i$ to global memory of $\mathbf{dK}$
    \EndFor
    
    \State Return $\mathbf{dQ}, \mathbf{dK}, \mathbf{dV}$
        
\end{algorithmic}
\end{algorithm}

The backward algorithm is presented in algorithm \ref{alg:bwd}. 
It closely mirrors the forward algorithm by storing the attention matrix in on-chip SRAM and dividing the query, key, and value matrices along the feature dimension.
The global memories of $\mathbf{V,O,dO,dQ,dK,dV}$ are accessed only once 
while the global memories of $\mathbf{Q,K}$ are accessed twice.
For on chip SRAM, the space complexity is also $\Theta(L^2+LC/r)$. More specifically,
the peak on chip memory usage is $2L^2 + 2LC/r$. For typical setting such as $L=64$ and $C/r=16$, 
the peak on chip memory usage is 41kb for fp32 data. The derivation of standard attention backward is represented in 
\cite{Dao2022FlashAttentionFA}. Algorithm \ref{alg:bwd} is 
different from it in two aspects: (1) there is no attention matrix $\mathbf{P}$ as input,
because we compute it on chip in forward pass; (2) tiling along feature dimension.

Note that algorithm \ref{alg:fwd} and \ref{alg:bwd} show the processing along sequence dimension and feature dimension. 
In real implementation, they are parallelized along head dimension (for multi head attention) and batch dimension.

\section{Benchmark}
We implement Flash Window Attention using Triton \cite{Tillet2019TritonAI} and PyTorch. 
Specifically, we develop GPU kernels for algorithms \ref{alg:fwd} and \ref{alg:bwd} using Triton and 
integrate them into PyTorch as an autograd function.
All experiments are conducted on an NVIDIA GeForce RTX 4090 GPU. For these experiments, 
the number of chunks along the feature dimension is set to $r=C/16$

\subsection{Attention Computation}

\begin{figure}[t]
    \centering
    \begin{subfigure}[b]{0.45\textwidth}
        \centering
        \includegraphics[width=\textwidth]{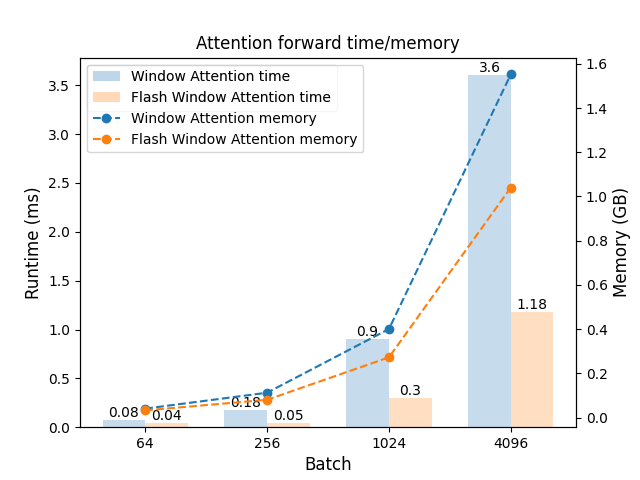}
    \end{subfigure}
    \hfill
    \begin{subfigure}[b]{0.45\textwidth}
        \centering
        \includegraphics[width=\textwidth]{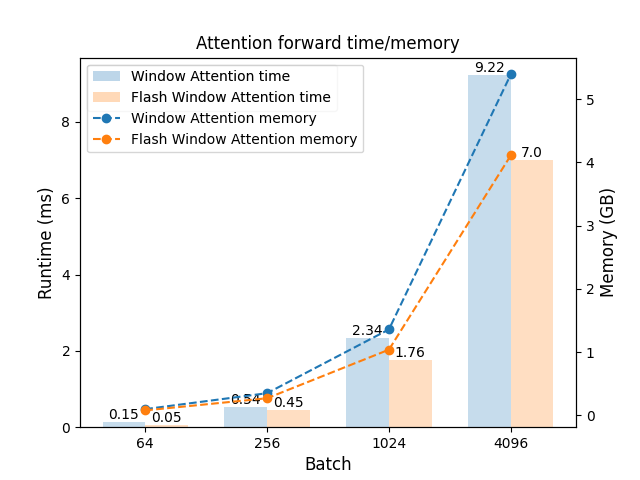}
    \end{subfigure}
    \caption{Comparison of forward attention computation. $C=64$ for left figure and $C=256$ for right figure. 
    Bars for running time while lines for memory usage. Note that $batch$ means the number of sequences after window partition, 
    instead of the batch size.}
    \label{fig:fwd1}
\end{figure}
\begin{figure}[t]
    \centering
    \begin{subfigure}[b]{0.45\textwidth}
        \centering
        \includegraphics[width=\textwidth]{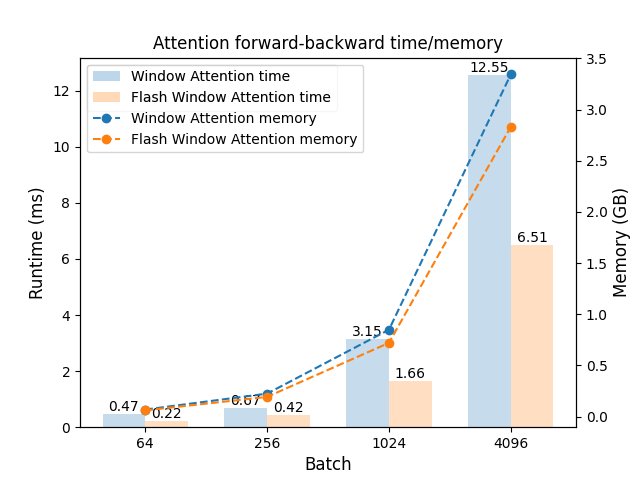}
    \end{subfigure}
    \hfill
    \begin{subfigure}[b]{0.45\textwidth}
        \centering
        \includegraphics[width=\textwidth]{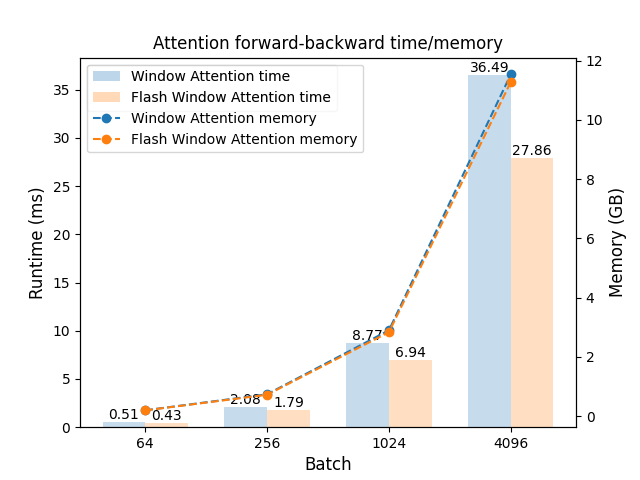}
    \end{subfigure}
    \caption{Comparison of forward-backward attention computation. $C=64$ for left figure and $C=256$ for right figure. 
    Bars for running time while lines for memory usage.}
    \label{fig:bwd1}

\end{figure}
We evaluate the efficiency of multi-head attention computation by comparing standard window attention with Flash Window Attention. 
The window attention baseline is implemented as in Swin Transformer \cite{Liu2021SwinTH}. 
The input to the attention mechanism consists of window-partitioned query, key, and value tensors with a shape of
$Batch \times head \times L \times C$.
We fix the number of $head$ to 4 and the length of sequence $L$ to 64. 
The forward performance is shown in figure \ref{fig:fwd1}. 
Our method achieves up to 300\% speedup. And the memory usage is less than the original window attention.
The forward-backward performance is shown in figure \ref{fig:bwd1}. 
Our method is still better in terms of both running time and memory usage. 
When $C$ is increased from 64 to 256, the performance gap is shrunk. 
The reason is that the feature dimension is processed chunk by chunk. So the degree of parallelism along this dimension 
is not sufficient. In \cite{Liu2021SwinTH}, $C$ is set to 32.

\subsection{End-to-End Running}

\begin{figure}
    \centering
    \begin{subfigure}[b]{0.45\textwidth}
        \centering
        \includegraphics[width=\textwidth]{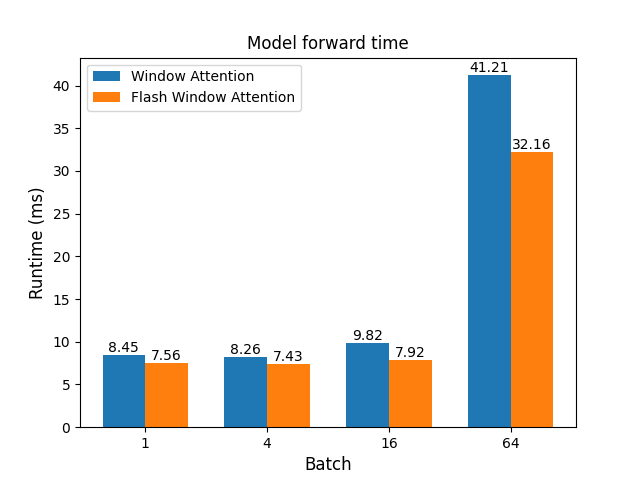}
    \end{subfigure}
    \hfill
    \begin{subfigure}[b]{0.45\textwidth}
        \centering
        \includegraphics[width=\textwidth]{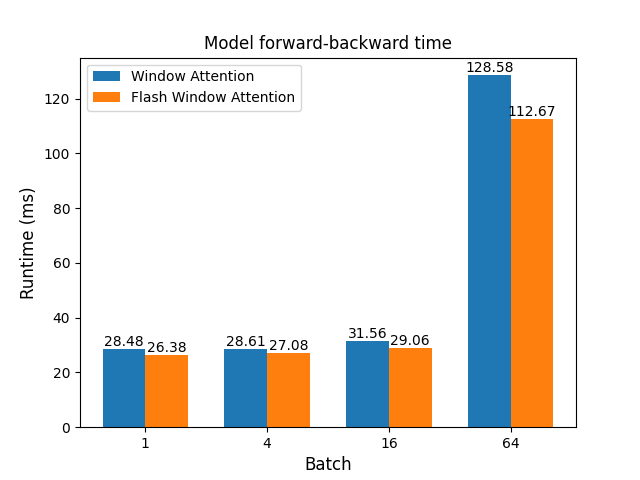}
    \end{subfigure}
    \caption{Comparison of end-to-end running of Swin Transformer.}
    \label{fig:end}

\end{figure}

We evaluate the end-to-end running time of Swin Transformer with window attention and Flash Window Attention. 
All settings are the same as the original paper, i.e. window size is $7\times7$ and input size is $224\times224$. 
As seen in figure \ref{fig:end}, our method achieves at least 10\% of end-to-end speedup. The speedup is more significant 
for larger image batch size. We don't find the significant difference of memory usage.

\begin{figure}[t]
    \centering
    \begin{subfigure}[b]{0.45\textwidth}
        \centering
        \includegraphics[width=\textwidth]{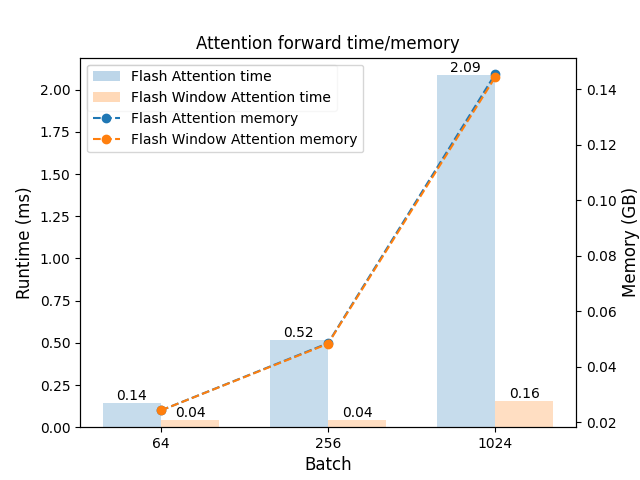}
    \end{subfigure}
    \hfill
    \begin{subfigure}[b]{0.45\textwidth}
        \centering
        \includegraphics[width=\textwidth]{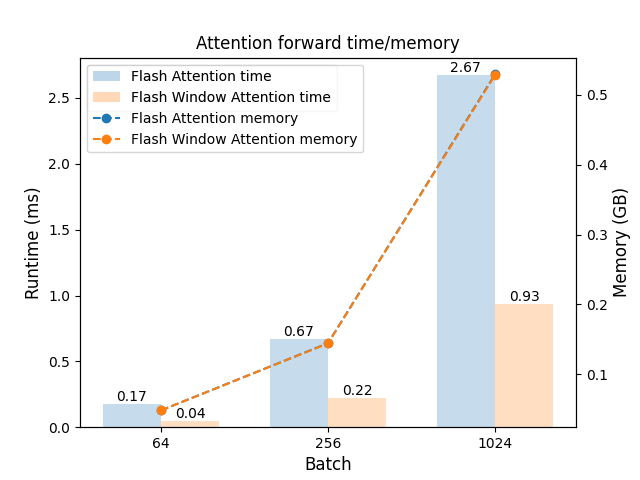}
    \end{subfigure}
    \caption{Comparison of forward attention computation. $C=64$ for left figure and $C=256$ for right figure. 
    Bars for running time while lines for memory usage. Note that $batch$ means the number of sequences after window partition, 
    instead of the batch size.}
    \label{fig:fwd3}
\end{figure}

\begin{figure}[!t]
    \centering
    \begin{subfigure}[b]{0.45\textwidth}
        \centering
        \includegraphics[width=\textwidth]{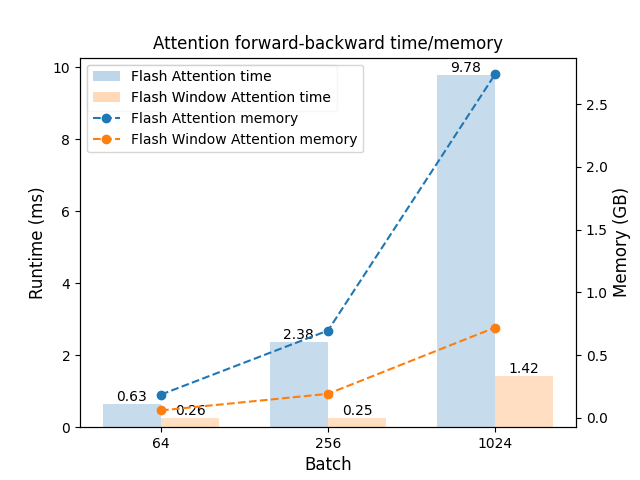}
    \end{subfigure}
    \hfill
    \begin{subfigure}[b]{0.45\textwidth}
        \centering
        \includegraphics[width=\textwidth]{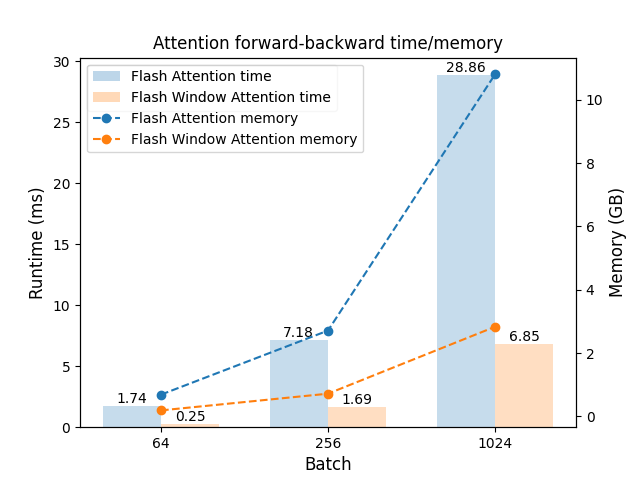}
    \end{subfigure}
    \caption{Comparison of forward-backward attention computation. $C=64$ for left figure and $C=256$ for right figure. 
    Bars for running time while lines for memory usage.}
    \label{fig:bwd3}

\end{figure}

\subsection{Compare with Flash Attention}

We evaluate the efficiency of multi-head attention computation by comparing flash attention \cite{Dao2022FlashAttentionFA,Dao2023FlashAttention2FA} 
with Flash Window Attention. The flash attention baseline is implemented as in \url{github.com/Dao-AILab/flash-attention}. 
Since flash attention doesn't support fp32 format, we use fp16 format for fair comparison. As see in figure \ref{fig:fwd3} and \ref{fig:bwd3}, 
flash attention is much slower than our method. As expected, the memory usage of forward pass is the same. 
However, flash attention requires much more memory for backward pass. When $batch=4096$, out of memory occurs. 

\section{Discussion}
In this report, we adopt the flash scheme for window attention and proposed Flash Window Attention, 
based on the following two observations:
\begin{itemize}
    \item For short sequences, the entire attention matrix can be stored on chip SRAM. 
    \item The computation of attention is decomposable along feature dimension.
\end{itemize}
In typical settings, we achieve up to 300\% speedup of attention computation and 30\% speedup of end-to-end running.
The limitation of is that we can't deal with very large window size such as $32 \times 32$ (out of on chip memory). 
In this case, the original flash attention is more suitable. 
In the future, we will extend our method to more general window patterns.

\bibliographystyle{plainnat}
\bibliography{main.bib}

\end{document}